\documentclass[lettersize,twoside, journal]{IEEEtran}

\usepackage{graphics}           
\usepackage{times}              
\usepackage{amsmath}            
\usepackage{amssymb}            
\usepackage{graphicx}
\usepackage{algorithm}
\usepackage[noend]{algpseudocode}
\usepackage{booktabs}
\usepackage{color}
\usepackage[nocompress]{cite} 
\definecolor{instructioncolor}{rgb}{.5,.5,.5}

\usepackage[font=small]{caption}

\def\eqref#1{Eq.~(\ref{#1})}

\makeatletter
\usepackage{xspace}
\DeclareRobustCommand\onedot{\futurelet\@let@token\@onedot}
\def\@onedot{\ifx\@let@token.\else.\null\fi\xspace}

\makeatother

\usepackage{array}
\newcolumntype{L}[1]{>{\raggedright\let\newline\\\arraybackslash\hspace{0pt}}m{#1}}
\newcolumntype{C}[1]{>{\centering\let\newline\\\arraybackslash\hspace{0pt}}m{#1}}
\newcolumntype{R}[1]{>{\raggedleft\let\newline\\\arraybackslash\hspace{0pt}}m{#1}}

\usepackage[T1]{fontenc}
\usepackage{aecompl}
\usepackage{flushend}
\usepackage{dsfont}
\usepackage{graphicx}
\usepackage[colorlinks=true,
            citecolor=blue,
            linkcolor=blue,
            anchorcolor=blue,
            urlcolor=blue]{hyperref}
\usepackage{amsmath,amsfonts,amssymb}
\usepackage{algorithm}
\usepackage{bbm}
\usepackage{multirow}
\usepackage{cite}
\usepackage{url}
\usepackage{textcomp}
\usepackage{stfloats}
\usepackage{verbatim}
\usepackage[table,xcdraw]{xcolor}
\usepackage{float}
\usepackage{lineno}

\makeatletter
\renewcommand{\maketag@@@}[1]{\hbox{\m@th\small\normalfont#1}}%
\makeatother

\newcommand{\keywordss}[1]{\par\textbf{\textit{Index Terms---}}\textbf{#1}}

\title{CoFiI2P: Coarse-to-Fine Correspondences-Based Image to Point Cloud Registration}

\author{
Shuhao Kang\textsuperscript{*}, 
Youqi Liao\textsuperscript{*}, 
Jianping Li\textsuperscript{\textdagger},~\IEEEmembership{Member,~IEEE}, 
Fuxun Liang,
Yuhao Li,
Xianghong Zou, \\
Fangning Li,
Xieyuanli Chen,
Zhen Dong,~\IEEEmembership{Member,~IEEE}, 
Bisheng Yang
\thanks{
Manuscript received: May. 14, 2024; Revised: Aug. 8, 2024; Accepted: Sep. 9, 2024. This paper was recommended for publication by
  Editor Cesar Cadena Lerma upon evaluation of the Associate Editor and Reviewers' comments. Digital Object Identifier (DOI): see top of this page. 
  
This study was jointly supported by the National Natural Science Foundation Project (No. 42201477, No. 42130105), the Open Fund of Hubei Luojia Laboratory (No. 2201000054) and Open Fund of Key Laboratory of Urban Spatial Information, Ministry of Natural Resources (Grant No. 2023ZD001). (Shuhao Kang and Youqi Liao are co-first authors and contribute equally to the paper) (Corresponding author: Jianping Li)}  
        
\thanks{
S. Kang is with the Technical University of Munich, Germany. Y. Liao is with the  Wuhan University and Hubei Luojia Laboratory, China. Z. Dong, F. Liang, Y. Li, X. Zou and B. Yang are with the Wuhan University, China. J. Li is with the Nanyang Technological University, Singapore. F. Li is with the Beijing Urban Construction Exploration and Surveying Design Research Institute Co. Ltd. X. Chen is with the National University of Defense Technology, China.}
}

\begin{document}
\maketitle
\markboth{IEEE Robotics and Automation Letters}%
{S. Kang, \MakeLowercase{\textit{et al.}}: CoFiI2P: Coarse-to-Fine Correspondences-Based Image to Point Cloud Registration}

\setlength{\textfloatsep}{1.3em}
\setlength{\dbltextfloatsep}{1.3em}

\begin{abstract}
Image-to-point cloud (I2P) registration is a fundamental task for robots and autonomous vehicles to achieve cross-modality data fusion and localization. Current I2P registration methods primarily focus on estimating correspondences at the point or pixel level, often neglecting global alignment. As a result, I2P matching can easily converge to a local optimum if it lacks high-level guidance from global constraints. To improve the success rate and general robustness, this paper introduces CoFiI2P, a novel I2P registration network that extracts correspondences in a coarse-to-fine manner. First, the image and point cloud data are processed through a two-stream encoder-decoder network for hierarchical feature extraction. Second, a coarse-to-fine matching module is designed to leverage these features and establish robust feature correspondences. Specifically, in the coarse matching phase, a novel I2P transformer module is employed to capture both homogeneous and heterogeneous global information from the image and point cloud data. This enables the estimation of coarse super-point/super-pixel matching pairs with discriminative descriptors. In the fine matching module, point/pixel pairs are established with the guidance of super-point/super-pixel correspondences. Finally, based on matching pairs, the transformation matrix is estimated with the EPnP-RANSAC algorithm. Experiments conducted on the KITTI Odometry dataset demonstrate that CoFiI2P achieves impressive results, with a relative rotation error (RRE) of 1.14 degrees and a relative translation error (RTE) of 0.29 meters, while maintaining real-time speed. These results represent a significant improvement of 84\% in RRE and 89\% in RTE compared to the current state-of-the-art (SOTA) method. Additional experiments on the Nuscenes dataset confirm our method's generalizability.
The project page is available at \url{https://whu-usi3dv.github.io/CoFiI2P}. 
\end{abstract}

\keywordss{ Image-to-Point Cloud Registration, Coarse-to-Fine Correspondences, Transformer Network}

\begin{figure}[!t]
\centering
\includegraphics[width=3.0in]{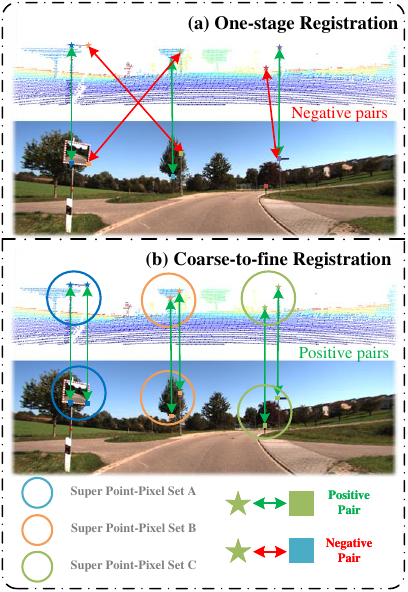}
\caption{Comparison of existing one-stage I2P registration and proposed coarse-to-fine I2P registration. (a) The existing one-stage registration pipeline. The matching pairs are directly established at the point/pixel level, leading to a significant number of mismatches. (b) Our coarse-to-fine matching pipeline. Under the guidance of super point-to-pixel pairs, point-to-pixel pairs are generated from the existing super pairs, which effectively eliminates most mismatches.}  
\vspace{-0.2cm}
\label{fig_1}
\end{figure}

\section{Introduction}\label{section:1}
Estimating the six degrees of freedom (6-DoF) pose of a monocular image relative to a pre-built point cloud map is a fundamental requirement for robots and autonomous vehicles ~\cite{wang2023camo, li2024hcto, li2018automatic}. Due to the limited onboard resources, robots are often equipped with only a monocular camera while facing challenges related to scale ambiguity in absolute localization and depth sensing. Establishing an accurate pose transformation between the image coordinate system and the pre-built point cloud coordinate system is crucial. This transformation not only precisely localizes the robot but also effectively reduces the scale uncertainty inherent in the monocular data~\cite{yan20222dpass}.

However, cross-modality registration has its inherent challenges. Some existing methods use hand-crafted detectors and descriptors for I2P registration~\cite{zhang2021line,yuan2021pixel}.  These approaches rely on structured features like edge features~\cite{yuan2021pixel}, which are limited by specific environmental conditions. With the rapid development of deep learning (DL), learning-based I2P registration approaches~\cite{feng20192d3d, li2021deepi2p, ren2022corri2p} have been proposed to extract representative keypoints and descriptors. 2D3D-MatchNet~\cite{feng20192d3d} proposed a novel Siamese network to learn the cross-modality descriptor but the manually designed detectors matched poorly. To alleviate the difficulty of correspondences construction and improve the registration success rate, DeepI2P~\cite{li2021deepi2p} converts the registration problem to the classification problem. A novel binary classification network is designed to distinguish whether the projected points are within or beyond the camera frustum. The classification results are passed into an inverse camera projection solver to estimate the transformation between the camera and laser scanners. As a large number of points on the boundary are misclassified, the accuracy of the camera pose is still limited. CorrI2P~\cite{ren2022corri2p} proposed an overlap region detector for both image and point cloud, then pixels and points in the overlap region are matched to obtain I2P correspondences. Feature fusion module is exploited to fuse the point cloud and image information. Although CorrI2P~\cite{ren2022corri2p} has significant improvement over DeepI2P~\cite{li2021deepi2p}, matching merely on one stage, namely, the pixel-point level without global alignment guidance, can lead to local minima and instability. 

Inspired by recent coarse-to-fine matching schedules and transformers in image-to-image (I2I) registration~\cite{sun2021loftr,jiang2021cotr} and point cloud-to-point cloud (P2P) registration approaches~\cite{qin2022geometric}, this paper proposes the \textbf{Co}arse-to-\textbf{Fi}ne \textbf{I}mage-\textbf{to}-\textbf{P}oint cloud (CoFiI2P) network for I2P registration. The I2P transformer with self- and cross-attention modules is embedded into the network for global alignment.
Overall, the main contributions of this work are as follows. 
\begin{enumerate}
  \item A novel coarse-to-fine I2P registration network is proposed to align image and point cloud in a progressive way. The coarse matching step provides rough but robust super-point/super-pixel correspondences for the following fine matching step, which filters out most mismatched pairs and reduces the computation burden. The fine matching step achieves accurate and reliable point/pixel correspondences with the global guidance.
  \item A novel I2P transformer that incorporates both self-attention and cross-attention modules is proposed to enhance its global-aware capabilities in homogeneous and heterogeneous data. The self-attention module enables the capture of spatial context within the same modality data, while the cross-attention module facilitates the extraction of hybrid features from both the image and point cloud data.
\end{enumerate}

\section{Related work}\label{section:2}

\subsection{Same-modality Registration}
\subsubsection{I2I Registration} Before the age of deep-learning, hand-crafted detectors and descriptors (i.e., SIFT~\cite{lowe2004distinctive}  and ORB~\cite{rublee2011orb}) are widely used to extract correspondences for matching. Compared to traditional methods, learning-based methods improve the robustness and accuracy of image matching with large viewpoint differences and illumination changes. 
SuperPoint~\cite{detone2018superpoint} proposed a self-learning training method through homography adaptation. SuperGlue~\cite{sarlin2020superglue} proposed an attention-based graph neural network (GNN) for feature matching. Patch2Pix~\cite{zhou2021patch2pix} is the first work to obtain patch matches and regress pixel-wise matches in a coarse-to-fine manner. 
 
\subsubsection{P2P Registration} Point cloud registration aims to estimate the optimal rigid transformation of two point clouds. Correspondence-based methods~\cite{deng2018ppf,gojcic2019perfect} estimate the correspondences first and recover the transformation with robust estimation methods~\cite{fischler1981random}. RoReg~\cite{wang2023roreg} embeds orientation information of point cloud to estimate local orientation and refine coarse rotation through residual regression to achieve fine registration. In the transformer era, point-based transformers~\cite{yu2021cofinet,qin2022geometric} have emerged and shown great performance. CoFiNet~\cite{yu2021cofinet} followed LoFTR's~\cite{sun2021loftr} design and proposed the coarse-to-fine correspondences for registration, which computed coarse matches with descriptors strengthened by the transformer and refined coarse matches through density adaptive matching module.  GeoTransformer~\cite{qin2022geometric} uses the Transformer to fully exploit the 3D properties of point clouds. However, I2I or P2P registration methods can not be directly used for the I2P registration, which extracts heterogeneous features from cross-modality data.

\subsection{Cross-modality Registration}
To address the cross-modality registration problem, a variety of I2P registration methods have been proposed, which could be roughly divided into two categories: I2P fine registration (initial transformation dependent) and I2P coarse registration (initial transformation free). The I2P fine registration methods~\cite{levinson2013automatic,zhang2021line,liao2023se} rely on initial transform parameters and are widely applied in sensor calibration. Although this paper focuses on the second category, namely coarse I2P registration without any initial transformation knowledge, we provide a comprehensive review of registration methods in both categories.
 
\subsubsection{Fine Registration Methods} Fine registration methods have been thoroughly studied for several decades. Early-stage works~\cite{zhang2004extrinsic,scaramuzza2007extrinsic} utilize various artificial targets as calibration constraints. In recent years, some approaches have argued that structured features shared in images and point cloud could be used for target-less I2P registration, e.g., edge feature~\cite{levinson2013automatic} and line feature~\cite{zhang2021line}.~\cite{levinson2013automatic} employs edge information and optimizes transformation according to the response value.~\cite{zhang2021line} extracts lines from image and point clouds for registration. Recently, researches on 2D and 3D semantic segmentation motivate semantic feature-based I2P registration. Se-calib~\cite{liao2023se} further studies semantic edges in I2P registration to employ more common semantic information instead of a certain type of object. Overall, fine registration methods achieve high accuracy but depend on the quality of the initial values.

\subsubsection{Coarse Registration Methods} 2D3D-Matchnet~\cite{feng20192d3d} is the pioneering method to regress the relative transform parameters with CNN. It extracts SIFT~\cite{lowe2004distinctive} and ISS~\cite{zhong2009intrinsic} keypoints from image and point cloud and learns descriptors with a Siamese network. Experimental results show that hand-crafted detectors from different modalities match poorly. HAS-Net~\cite{lai2021learning} proposed a novel network for learning cross-domain descriptors from 2D image patches and 3D point cloud volumes. However, both 2D3D-Matchnet and HAS-Net split the image and point cloud into patches and volumes, and then match with the descriptors of patches and volumes, resulting in the loss of long-range context and high-level information. DeepI2P~\cite{li2021deepi2p} proposed a feature fusion module to merge image and point cloud information and classified points in/beyond the camera frustum. CorrI2P~\cite{ren2022corri2p} predicted pixels and points in overlapping areas and matched with dense per-pixel/per-point features directly to get I2P correspondences. Although overlapping region detectors significantly reduce the number of false candidates, I2P registration only using the low-level feature without global guidance leads to serious mismatches. Inspired by the coarse-to-fine strategy in CoFiNet~\cite{yu2021cofinet}, we propose the CoFiI2P network, a coarse-to-fine I2P registration approach that integrates high-level correspondence information into low-level matching to effectively reject mismatches.
 
\section{Methodology}\label{section:3}

\begin{figure*}[!t]
\centering
\includegraphics[width=\textwidth]{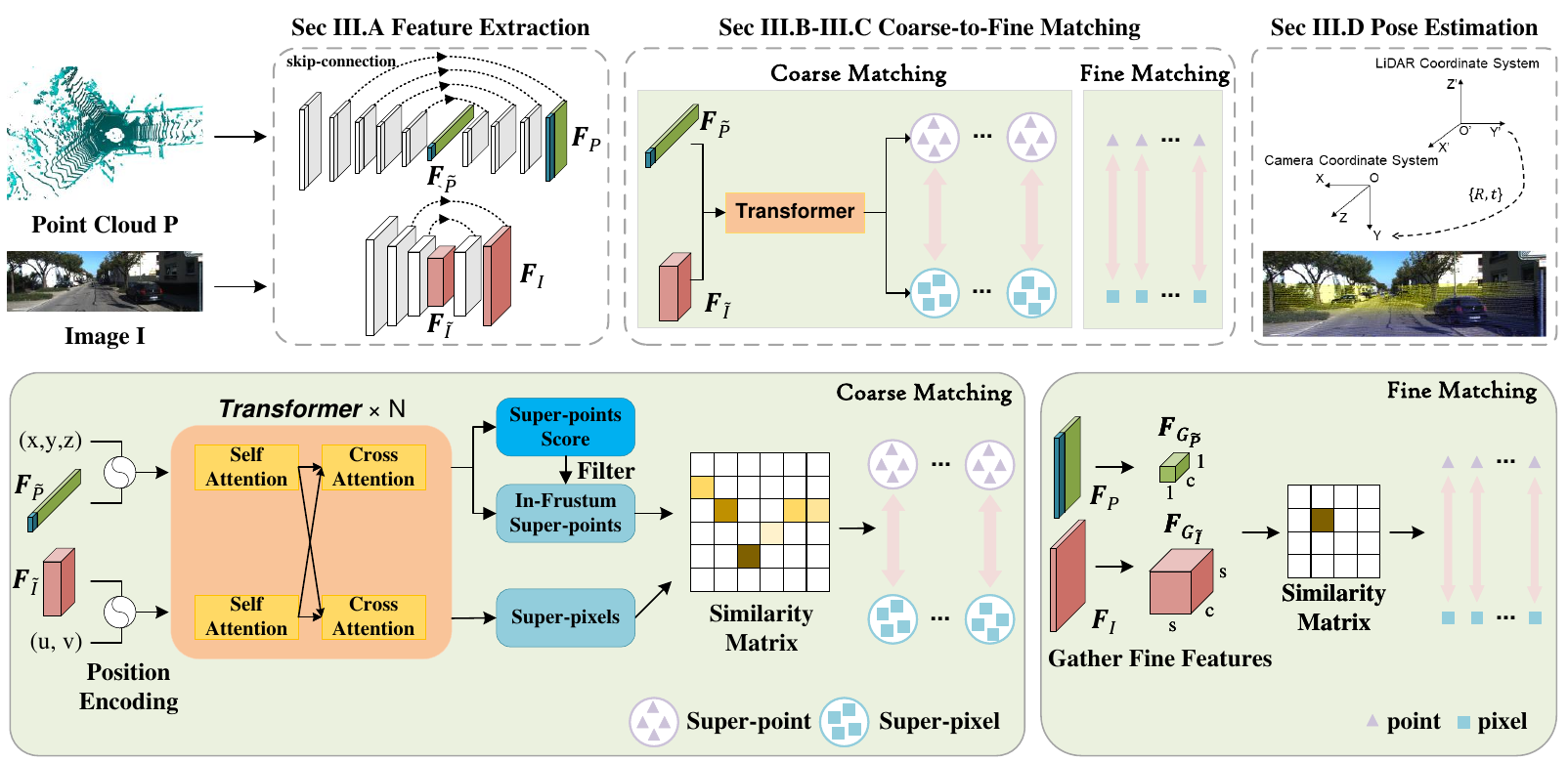}
\caption{Workflow of CoFiI2P. The proposed method consists of feature extraction, coarse matching, fine matching and pose estimation modules. Image and point cloud are sent to the feature extraction module to obtain coarse-level features and fine-level features (rendered in red for image and green for point cloud, respectively). The coarse-level features are strengthened by I2P transformer module and then matched with the cosine similarity. Fine features are gathered from the last layer of the decoder. In each super-point/super-pixel pair, the node point is set as the candidate and the corresponding pixel is selected from the super-pixel area, a $ w \times w $ window. The generated fine-level matching pairs are utilized to estimate the pose with the EPnP-RANSAC~\cite{lepetit2009ep,fischler1981random} algorithm.}  
\label{fig_2} 
\vspace{-0.1cm}
\end{figure*}

For convenient description, a pair of partially overlapped image and point cloud are defined as  $ \mathrm{I} \in \mathbb{R}^{W\times H \times 3}$ and  $ \mathrm{P} \in \mathbb{R}^{N \times 3}$, where $W$ and $H$ are the width and height, and $N$ is the number of points. The purpose of I2P registration is to estimate the relative transformation between the image $\mathrm{I}$ and point cloud $\mathrm{P}$, defined by a rotation matrix $\mathbf{R} \in \mathbb{R}^{3\times3}$ and a translation vector $\mathbf{t} \in \mathbb{R}^3$.

Our method adopts the coarse-to-fine manner to find the correct correspondences set $\Omega(\mathrm{I},\mathrm{P})$  and calculate the relative pose using EPnP-RANSAC~\cite{lepetit2009ep}. The CoFiI2P mainly consists of four modules: feature extraction (FE), coarse matching (CM), fine matching (FM), and pose estimation (PE). FE is an encoder-decoder structure network, that encodes raw inputs from different modalities into shared feature space and finds in-frustum super-points. CM and FM are cascaded two-stage matching modules. CM constructs coarse matching pairs at super-pixel/super-point level, and them FM constructs fine matching pairs at pixel/point level sequentially with the guidance of super-pixel/super-point correspondences. Lastly, the PE module exploits point-pixel matching pairs to regress the relative pose with the EPnP-RANSAC algorithm. The workflow of the proposed method is shown in Fig. \ref{fig_2}.    

\vspace{-0.1cm}
\subsection{Feature Extraction}
We utilize ResNet-34~\cite{he2016deep} and KPConv-FPN~\cite{thomas2019kpconv} as the backbones for image and point cloud to extract multi-level features. The encoder progressively embeds raw inputs into high-dimensional features, and the decoder propagates high-level features to low-level features with skip-connection for per-pixel/point feature generation. Specifically, points and pixels at the coarsest resolution ($\frac{1}{8}$ of original resolution) ${\tilde{P}}$ and ${\tilde{I}}$ are treated as superpoints and superpixels for coarse matching, and ${P}$ and ${I}$ at the finest resolution ($\frac{1}{2}$ of original resolution) are used for fine matching. We use $\boldsymbol{F}_{\tilde{P}}$ and $\boldsymbol{I}_{\tilde{P}}$ to denote coarse-level features, and $\boldsymbol{F}_{P}$ and $\boldsymbol{I}_{P}$ for fine-level features. For each superpoint, we construct the local points group $G_{\tilde{p}}$ with point-to-node strategy in the geometric space: 
    \begin{equation}
       G_{\tilde{p}}=\{p \in P \lvert  \| p - \tilde{p} \| < r_g \},
    \end{equation}
    where $r_g$ is the chosen radius. For the superpixels, we first locate their positions on the fine-level feature map, and then crop sets of local pixel patches with window size $w \times w$.

\vspace{-0.1cm}
\subsection{I2P Coarse Matching}
In the CM module, I2P transformer is utilized to capture the geometric and spatial consistency between image and point cloud. Each stage of the I2P transformer consists of a self-attention block for inter-modality long-range context and a cross-attention module for intra-modality feature exchange. The self-attention and cross-attention modules are repeated several times to extract well-mixed features for super-point/super-pixel correspondence matching.

 \subsubsection{I2P Transformer}~\cite{zhang2022topformer,wan2023seaformer} have shown that vision transformer (ViT) outperforms traditional CNN-based methods with a large margin in classification, detection, segmentation and other downstream tasks. Furthermore, recent approaches~\cite{jiang2021cotr,sun2021loftr,yu2023rotation} have introduced transformer modules for I2I and P2P registration tasks. Therefore, we introduce the I2P transformer module customized for cross-modality registration task to enhance the representability and robustness of descriptors. Different from ViT used in the same-modality registration tasks, our I2P transformer contains both self-attention modules for space context capturing in homogeneous data and cross-attention modules for hybrid feature integration among heterogeneous data. 
 
 For the self-attention module, given a coarse-level feature map $\mathbf{F}\in \mathbb{R}^{N \times C}$ of image or point cloud, the query, key and value vectors $\mathbf{q}, \mathbf{k},\mathbf{v}$ are generated as:
    \begin{equation}
        \mathbf{q}=\mathbf{W_qF},\mathbf{k}=\mathbf{W_kF},\mathbf{v}=\mathbf{W_vF},
    \end{equation}
where $\mathbf{W}_q,\mathbf{W}_k \in \mathbb{R}^{C_{qk}^{sa}\times C},\mathbf{W}_v \in \mathbb{R}^{C_v^{sa} \times C}$ are learnable weight matrixs. Then, the global attention enhanced feature map $\mathbf{F}^{sa}$  is calculated as: 
\begin{equation}\label{eq.3} 
\mathbf{F}^{sa}=\mathrm{softmax}(\frac{\mathbf{q}\mathbf{k}^{\top}}{\sqrt{C}})\mathbf{v}.
\end{equation}
Softmax operates row-wise on the attention matrix $\mathbf{q}\mathbf{k}^{\top}$ to obtain the weights on the values. Extracted global-aware features $\mathbf{F}^{sa}$ are fed into the feed-forward network (FFN) to fuse the spatial relation information in channel dimension. Given a feature map $\mathbf{F}$, the relative positions are encoded with multi-layer perception (MLP)~\cite{wu2021rethinking}.

Cross-attention is designed for fusing image and point cloud features in the I2P registration task. Given the feature maps $\mathbf{F}_{\tilde{P}},\mathbf{F}_{\tilde{I}} $ for super-points set $\tilde{P}$ and super-pixels set $\tilde{I}$, cross-attention enhanced feature maps $\mathbf{F}_{\tilde{P}}^{ca}$ of point cloud and $\mathbf{F}_{\tilde{I}}^{ca}$ of image are calculated as:
\begin{equation}\label{eq:5}
   \begin{split}
      &\mathbf{F}_{\tilde{P}}^{ca}=\mathrm{softmax}(\frac{\mathbf{q}_{\tilde{P}}\mathbf{k}_{\tilde{I}}^{\top}}{\sqrt{C}})\mathbf{v}_{\tilde{P}},\\
      &\mathbf{F}_{\tilde{I}}^{ca}=\mathrm{softmax}(\frac{\mathbf{q}_{\tilde{I}}\mathbf{k}_{\tilde{P}}^{\top}}{\sqrt{C}})\mathbf{v}_{\tilde{I}}, 
   \end{split} 
\end{equation}
where $\mathbf{q}_{\tilde{P}},\mathbf{k}_{\tilde{P}},\mathbf{v}_{\tilde{P}}$ are query, key and value vectors of point cloud feature $\mathbf{F}_{\tilde{P}}$, and $\mathbf{q}_{\tilde{I}},\mathbf{k}_{\tilde{I}},\mathbf{v}_{\tilde{I}}$ are query, key and value vectors of image feature $\mathbf{F}_{\tilde{I}}$. The softmax operation is the same as in the self-attention module.

\textbf{Remark 1.} While the self-attention module encodes the spatial and geometric features for each super-pixel and super-point, the cross-attention module injects the geometric structure information and texture information across image and point cloud respectively. Outputs of the I2P transformer carry powerful cross-modality information for matching.

\subsubsection{Super-point/-pixel Matching} The field of view (FoV) of the monocular camera is obviously smaller than the laser scans of 3D Lidar (e.g., Velodyne-H64), which sweeps 360 degrees in the horizontal direction. Therefore, only a small number of super-points are in the camera frustum.  To filter out beyond-frustum super-points, we add a simple binary classification head to predict super-points in or beyond the frustum of the camera.  The coarse level correspondences are estimated between in-frustum super-points set $\tilde{P}$ and super-pixels set $\tilde{I}$ by 
finding the nearest super-pixel $\tilde{i}$ in feature space: 
\begin{small}
\begin{equation}
     \tilde{\Omega}_\mathcal{M}=\{(\tilde{p}_x \in \tilde{P}, \tilde{i}_y \in \tilde{I})  \lvert
     y =\mathop{\mathrm{argmin}}\limits_{\lvert \tilde{I} \rvert} \|  \mathbf{F}_{\tilde{P}}(\tilde{p}_x) - \mathbf{F}_{\tilde{I}}(\tilde{i}_y) \| \},
\end{equation}
\end{small}
where $\mathbf{F}_{\tilde{P}}$ and $\mathbf{F}_{\tilde{I}}$ are corresponding feature maps.

\vspace{-0.1cm}
\subsection{I2P Fine Matching}

 The first-stage matching at the coarse level constructs robust super-point/super-pixel pairs $\tilde{\Omega}_{\mathcal{M}}$ but leads to poor registration accuracy. To obtain high-quality I2P correspondences, we generate fine correspondences based on the coarse matching results. 
 In each super-point/super-pixel correspondence $(\tilde{p}_x,\tilde{i}_y)$, the super-point $\tilde{p}_x$ is expanded to a local points group $G_{\tilde{p}_x}$ and the super-pixel $\tilde{i}_y$ is expanded to a local pixels patch $G_{\tilde{i}_y}$. Considering the uneven geometric distribution of point cloud and computational efficiency, only the node points in local patch groups are selected to establish correspondences. For each node point $p_n$, we select the pixel $i_k\in G_{\tilde{i}_y}$ that lies nearest in the feature space. Point-pixel pairs in each super-point-to-super-pixel pair are stacked together as the fine corresponding pairs $\Omega_{\mathcal{M}}$. With point feature map $\mathbf{F}_{G_{\tilde{p}}}$  and corresponding pixel feature map $\mathbf{F}_{G_{\tilde{i}}}$ of local patch, the  fine matching process is defined as:  

\begin{small}
    \begin{equation}
\begin{split}
    \Omega_{\mathcal{M}}= \big\{(p_n \in G_{\tilde{p}}, i_k \in G_{\tilde{i}})  \lvert 
      k=\mathop{\mathrm{argmin}}\limits_{\lvert G_{\tilde{i}} \rvert} \|  \mathbf{F}_{G_{\tilde{p}}}(p_n) - \mathbf{F}_{G_{\tilde{i}}}(i_k) \| \big\}.
\end{split}
\end{equation} 
\end{small}

\subsection{EPnP-RANSAC based Pose Estimation}
	With the precise point-pixel pairs $\Omega_{\mathcal{M}}$, the relative transformation can be solved with EPnP~\cite{lepetit2009ep} algorithm. As mentioned in previous approaches, wrong matching may infiltrate into the point-pixel pairs and decrease the registration accuracy. In the CoFiI2P, the EPnP-RANSAC~\cite{lepetit2009ep,fischler1981random} algorithm is used for robustly estimating camera relative pose. 

 \vspace{-0.1cm}
\subsection{Loss Function}
	To learn the coarse level descriptors, fine level descriptors and in/beyond-frustum super-points classification simultaneously, we introduce a joint loss $\mathcal{L}$ consisting of coarse level descriptor loss $\mathcal{L}_{coarse}$ , fine level descriptor loss $\mathcal{L}_{fine}$ and classification loss $\mathcal{L}_{classify}$. The classification loss encourages the network to correctly label each super-point, and two descriptor losses pull positive matching pairs closer together and push negative matching pairs farther apart in the feature space. 
 
 The cosine similarity $s(p_{x},i_{y})$  of point cloud feature vector $\mathbf{F}_{p_x}$ and image feature vector $\mathbf{F}_{i_y}$ is defined as :
 \begin{equation}
     s(p_{x},i_{y}) = \frac{<{\mathbf{F}_{p_x}},{\mathbf{F}_{i_y}}>}{\| {\mathbf{F}_{p_x}} \| \| {\mathbf{F}_{i_y}} \|},
 \end{equation}
 and the distance $d(p_{x},i_{y})$ is defined as:
 \begin{equation}
     d(p_{x},i_{y}) = 1 - s(p_{x},i_{y}).
 \end{equation}
On the coarse level, the positive anchor $\tilde{i}_{pos}$ for each in-frustum super-point  $\tilde{p}_x$ is sampled from the ground-truth pairs set $\tilde{\Omega}_{\mathcal{M}^\star}$:
\begin{equation}
    \tilde{\Omega}_{\mathcal{M}^\star}=\{(\tilde{p}_x \in \tilde{P},\tilde{i}_{pos} \in \tilde{I}) \lvert \tilde{i}_{pos} = \Gamma(\mathbf{T}\tilde{p}_x) \},
\end{equation}
where $\mathbf{T}$ is the ground-truth transform matrix from the point cloud coordinate system to the image frustum coordinate system, and $\Gamma$ represents the mapping function that converts points from the camera frustum to the image plane coordinate system. The negative anchor $\tilde{i}_{neg}$ is defined as the super-pixel with the smallest distance to the $\tilde{p}_x$ in the feature space:
    
    \begin{equation} \label{eq_neg}
    \tilde{i}_{neg}=\mathop{\mathrm{argmin}}\limits_{\tilde{i}_{neg} \in \tilde{I}} {\| d(\mathbf{F}_{\tilde{p}_x},\mathbf{F}_{\tilde{i}_{neg}}) \|} \quad s.t. \quad \| \tilde{i}_{neg} - \tilde{i}_{pos} \| > r,
\end{equation}
where $r$ is the safe radius to remove adversarial samples.
Finally, with positive margin $\Delta_{pos}$ and negative margin $\Delta_{neg}$, coarse level descriptor loss is defined in a triplet way~\cite{lai2021learning} as Eq. (\ref{eq:12}) :
\begin{small}
\begin{equation}\label{eq:12}
\begin{split}
     \mathcal{L}_{coarse}=\sum_{(\tilde{p}_x,\tilde{i}_{pos},\tilde{i}_{neg})}\bigg(& \mathrm{max}\big(0,d(\tilde{p}_{x},\tilde{i}_{pos}) - \Delta_{pos}\big) +  \\
    & \mathrm{max}\big(0,\Delta_{neg} - d(\tilde{p}_{x},\tilde{i}_{neg})\big)\bigg).
\end{split}
\end{equation}
\end{small} 
Fine level descriptor loss is defined as modified circle loss~\cite{sun2020circle}. We randomly select $n$ in-frustum points and their positive anchor pixels and negative anchors, as Eq. (\ref{eq_neg}), then the descriptor loss is defined as:

\begin{small}
   \begin{equation}
\begin{split}
    \mathcal{L}_{fine} =  \log \bigg(
    & 1+ \exp\big({-\gamma\alpha_{pos}(s(p_x,i_{pos})- \Delta_{pos})}\big) \\
    & \sum_{(p_x,i_{pos},i_{neg})} \exp\big({\gamma {\alpha}_{neg} (p_x,s(p_x,i_{pos})- \Delta_{neg})}\big)\bigg),
\end{split}
\end{equation}
\end{small}
where $\alpha_{neg}$ and $\alpha_{pos}$ are the dynamic optimizing rates towards negative and positive pairs, and $\gamma$ is the scale factor. As in~\cite{sun2020circle}, the $\alpha_{neg}$ and $\alpha_{pos}$ are defined as:
\begin{equation}
\begin{split}
     &  \alpha_{neg}=\mathrm{max}\big(0,s(p_x,i_{neg})+\Delta_{neg}\big), \\
         &  \alpha_{pos}=\mathrm{max}\big(0,1+\Delta_{pos}-s(p_x,i_{pos})\big).
\end{split}
\end{equation}
Super-points classification is defined as a binary cross-entropy loss:
\begin{equation}
    \mathcal{L}_{classify}= - \sum_{\tilde{p}_x \in \tilde{P}} \big({\hat{s}_{\tilde{p}_x}} \log({s_{\tilde{p}_x}}) + (1-{\hat{s}_{\tilde{p}_x}})\log{(1-{s_{\tilde{p}_x}})}\big),
\end{equation}
To avoid ambiguity, we reuse $s_{\tilde{p}_x}$ for score prediction and $\hat{s}_{\tilde{p}_x}$ for ground truth label.
Overall, our loss function is 
\begin{small}
\begin{equation}
    \mathcal{L} = \lambda_1  \mathcal{L}_{coarse} + \lambda_2 \mathcal{L}_{fine} + \lambda_3 \mathcal{L}_{classify},
\end{equation}
\end{small}
where $\lambda_1,\lambda_2,\lambda_3$ are hyperparameters that control the weights between losses.

\vspace{-0.3cm}
\section{Experiments}\label{section:4}

\subsection{Experiment Setup}\label{section:4.1}

\subsubsection{Dataset} We evaluate our method on two public datasets, KITTI Odometry~\cite{geiger2012we} and Nuscenes~\cite{caesar2020nuscenes}.
\begin{itemize}
      \item KITTI Odometry~\cite{geiger2012we}: It consists of 11 sequences of images and point cloud data in urban environments with ground-truth calibration files. The camera intrinsic function $\Gamma$ extracted from calibration files is thought unbiased during experiments. Sequences 0-8 are used for training and 9-10 for testing as previous approaches~\cite{li2021deepi2p,ren2022corri2p}. The image resolution is resized to 160$\times$512 and the point cloud is downsampled and randomly selected to 20480.
    \item Nuscenes~\cite{caesar2020nuscenes}: The image and point cloud pairs are generated by the official SDK, where the point cloud is accumulated from the nearby frames and the image is from the current frame. We follow the official data split of nuScenes to utilize 850 scenes for training, and 150 scenes for testing. We downsample the image resolution to 160$\times$320 and the point cloud size to 20480 similarly.
\end{itemize}

\subsubsection{Baseline Methods} We compare the proposed CoFiI2P with two open-sourced I2P methods as follows.
\begin{itemize}
    \item DeepI2P~\cite{li2021deepi2p}: It uses the frustum classification and inverse camera projection to estimate the camera pose. We use officially released code for reimplementation and use the 2D and 3D inverse camera projection for optimization, namely DeepI2P (2D) and DeepI2P (3D).
    \item CorrI2P~\cite{ren2022corri2p}: It is the SOTA I2P method. CorrI2P predicts the overlapping area and establishes correspondences densely for pose estimation. We use officially released code for reimplementation. 
\end{itemize}

\subsubsection{Evaluation Metrics} We calculate relative rotation error (RRE), relative translation error (RTE) and registration recall (RR) to evaluate the registration accuracy. Inlier ratio (IR) used in previous approaches~\cite{yu2021cofinet,wang2021p2} is introduced to evaluate the quality of correspondences. For efficiency analysis, we report the frame-per-second (FPS) on the KITTI Odometry dataset, evaluated with a batch size of 1 on an Intel i9-13900K CPU and an NVIDIA RTX 4090 GPU. RRE and RTE are defined as :
\begin{equation}
\begin{split}
      \mathrm{RRE}=\sum_{i=1}^3 \lvert \mathbf{r}(i) \rvert, \quad
      \mathrm{RTE}=\| \mathbf{t}_{gt} - \mathbf{t}_e \|,
\end{split} 
\end{equation}
where $\mathbf{r}$ is the Euler angle vector of $\mathbf{R}_{gt}^{-1}\mathbf{R}_e$, $\mathbf{R}_{gt}$ and $\mathbf{t}_{gt}$ are the ground-truth rotation and translation matrix, $\mathbf{R}_e$ and $\mathbf{t}_e$ are the estimated rotation and translation matrix. 

RR metric estimates the percentage of correctly matched I2P pairs, indicating the descriptor learning ability of the network. RR is defined as:
\begin{equation}
    \mathrm{RR} = \mathds{1}( \mathrm{RRE}<\delta_r \quad \textit{and} \quad \mathrm{RTE} < \delta_t ) ,
\end{equation}
$(\delta_r,\delta_t)$ is the threshold to remove false registration results, i.e. $(10^\circ,5m)$. IR denotes the inlier ratio of matching pairs, which measures the accuracy of correspondences. The IR for point/pixel correspondences set $\Omega_{\mathcal{M}}$ is defined as :
\begin{equation}
    \mathrm{IR}_{\mathcal{M}}=\frac{1}{\lvert \Omega_{\mathcal{M}} \rvert} \sum_{(p_x,i_y)\in \Omega_{\mathcal{M}}} \mathds{1}(\|\Gamma(\mathbf{T}p_x) - i_y\| <\tau),
    \vspace{-0.1cm}
\end{equation}
in which $\tau$ is used to control the reprojection error tolerance. 

\subsubsection{Implementation Details}

With correct transform parameters provided by the calibration files during the training process, the ground truth correspondences $\Omega_\mathcal{M^\star}$ are established to supervise the network. We trained the whole network 25 epochs for the KITTI Odometry dataset and 10 epochs for the Nuscenes dataset. We use the Adam~\cite{kingma2014adam} to optimize the network, and the initial learning rate is 0.001 and multiple by 0.25 after every 5 epochs. For our joint loss, we set $\lambda_1=\lambda_2=\lambda_3=1$. The safe radius $r$, positive margin $\Delta_{pos}$ and negative margin $\Delta_{neg}$ in loss function are set to 1, 0.2 and 1.8. Scale factor $\gamma$ is set to 10. Model configurations and more implementation details are available in our open-source code.

\begin{table*}[t]
\renewcommand\arraystretch{1.1}
\footnotesize
\centering
\caption{Registration accuracy on the KITTI Odometry~\cite{geiger2012we} and Nuscenes~\cite{caesar2020nuscenes} datasets. The results are presented in the format of "mean $\pm$ standard deviation" among the test samples. $\uparrow$ means higher is better and $\downarrow$ means lower is better, respectively. The best results are highlighted in bold.
Results of DeepI2P on the KITTI Odometry dataset are taken from~\cite{li2021deepi2p} with identical settings.}
\label{tab:table1}
\begin{tabular}{cc|ccc|ccc|c}
\hline 
 \multirow{2}{*}{Method}       & \multirow{2}{*}{Threshold($^\circ/m$)}       & \multicolumn{3}{c|}{KITTI Odometry} & \multicolumn{3}{c|}{Nuscenes} & \multirow{2}{*}{FPS$\uparrow$} \\ \cline{3-8}
 &                             & RRE($^\circ$)$\downarrow$                  & RTE(m)$\downarrow$         & RR(\%) $\uparrow$ & RRE($^\circ$)$\downarrow$                          & RTE(m)$\downarrow$                           & RR(\%) $\uparrow$  & \\ \hline
DeepI2P (2D)~\cite{li2021deepi2p} &        10/5                    & $7.56 \pm 7.63$                         &       $3.28 \pm 3.09$                         &  -
&  $2.81 \pm 2.23$ & $1.48 \pm 0.88$  & 95.13 & 1.41 \\                         
DeepI2P (3D)~\cite{li2021deepi2p}                 & 10/5   &    $15.52 \pm 12.73$           & $3.17\pm3.22$           & -   & $6.60 \pm 19.33$  & $1.31 \pm 1.57$  & 38.64 & 0.71 \\ \hline
\multirow{3}{*}{CorrI2P~\cite{ren2022corri2p}}  & none/none                        & $6.93 \pm 29.03$ & $2.68 \pm 10.51$ & -   & $8.47 \pm 24.02$  & $5.30 \pm 8.10$  & - & \multirow{3}{*}{7.95} \\
                         & 45/10                      & $3.41\pm3.64$                      & $1.48 \pm 1.35$                      & 97.07 & $5.42 \pm 4.30$   &  $3.78 \pm 2.24$  & 91.07 \\
                         & 10/5                       & $2.70\pm1.97$                      & $1.24 \pm 0.87$                      & 90.66 &  $3.90 \pm 2.23$  & $2.61 \pm 1.23$   & 61.67 \\ \hline
\multirow{3}{*}{CoFiI2P} & none/none                        & $\boldsymbol{1.14} \pm \boldsymbol{0.78}$  & $\boldsymbol{0.29} \pm \boldsymbol{0.19}$  & -   
&  $\boldsymbol{2.04} \pm \boldsymbol{5.54}$  & $\boldsymbol{0.95} \pm \boldsymbol{5.04}$   & - & \multirow{3}{*}{\textbf{15.43}} \\ 
                         & 45/10                      & $\boldsymbol{1.14} \pm \boldsymbol{0.78}$                      & $\boldsymbol{0.29} \pm \boldsymbol{0.19}$                      & \textbf{100.00} & $\boldsymbol{1.88} \pm \boldsymbol{1.77}$   &  $\boldsymbol{0.81} \pm \boldsymbol{0.59}$  & \textbf{99.84} \\
                         & 10/5                       & $\boldsymbol{1.14} \pm \boldsymbol{0.78}$                      & $\boldsymbol{0.29} \pm \boldsymbol{0.19}$                      & \textbf{100.00} & $\boldsymbol{1.79} \pm \boldsymbol{1.22}$  & $ \boldsymbol{0.79} \pm \boldsymbol{0.53}$   & \textbf{99.18} \\ \hline
\end{tabular}
\end{table*}

\subsection{Registration Accuracy}

\begin{figure*}[!ht]
\centering
\includegraphics[width=7in]{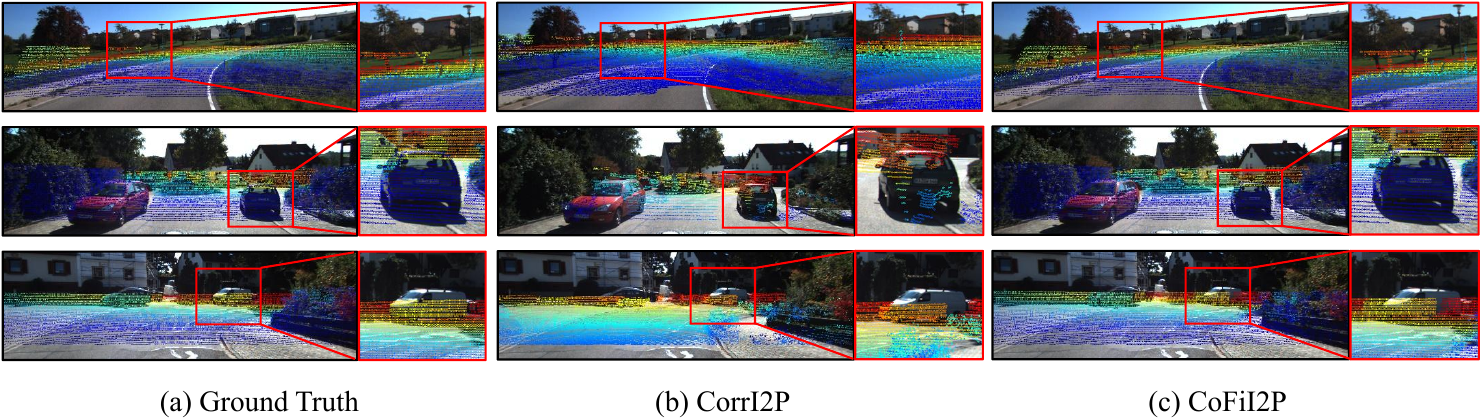}
\caption{Quantitative registration results on the KITTI Odometry dataset. The colors are rendered based on depth, ranging from blue in the foreground to red in the distance.}
\label{fig_5}
\end{figure*}

 We report the RRE and RTE as evaluation metrics under three different settings in Table \ref{tab:table1}, where $none/none$ means no specific thresholds for filtering out false registration frames and  $10^\circ/5m$ and $45^\circ/10m$ mean that any frames with registration errors exceeding these specified thresholds are ignored during the evaluation process. As shown in Table \ref{tab:table1}, the proposed method outperforms all baseline methods on the RRE and RTE registration metrics. Notably, our method achieves 100\% RR and 99.18\% RR under the hardest pair of thresholds $10^\circ/5m$ on the KITTI Odometry and Nuscenes datasets respectively,  which indicates that our method constructs robust correspondences and achieves accurate registration performance in most of evaluation scenes. Besides, we project the points to the image plane with transform parameters provided by ground-truth files, CorrI2P~\cite{ren2022corri2p} and CoFiI2P respectively and deploy the qualitative registration results in Fig. \ref{fig_5}. It shows that the CoFiI2P remains stable and provides more accurate results, which confirms our claim. Although our method is slightly slower during the model inference step, the pose estimation is extremely fast due to the small quantity and high quality of matching pairs, thereby still maintaining online speed.

 \begin{figure}[t]
\centering
\includegraphics[width=3.5in]{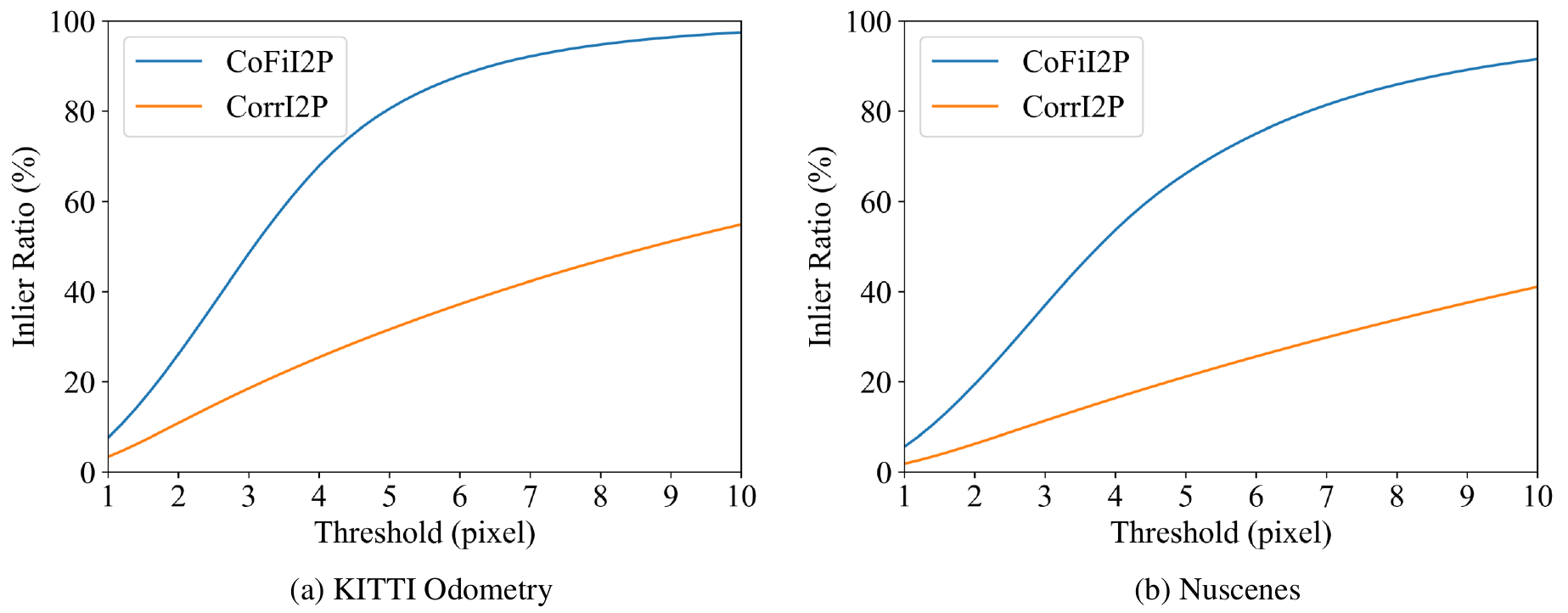}  
\caption{Quantitative results of correspondences. (a) and (b) shows the inlier ratio of our method (blue line) and CorrI2P (orange line) on the KITTI Odometry and Nuscenes dataset respectively.}
\label{fig_IR}
\vspace{0.6cm}
\centering
\includegraphics[width=3.5in]{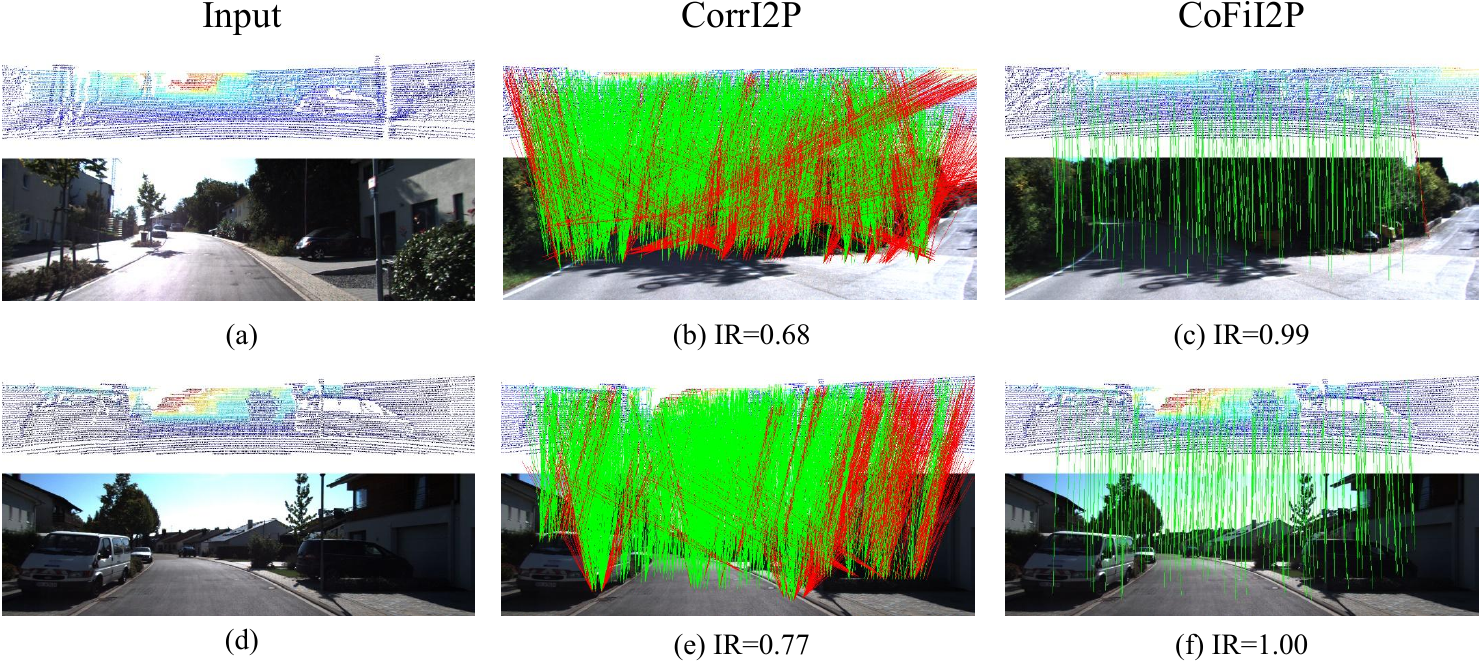} 
\caption{Qualitative results of correspondences on the KITTI Odometry \cite{geiger2012we} dataset. The first column shows the input, and the second and third column shows the correspondences of CorrI2P and our CoFiI2P respectively. The green lines represent correct matches, while the red lines represent incorrect matches.}
\label{fig_4}
\end{figure}

For the super-point frustum classification, We introduce the IR metrics to evaluate the quality of the correspondence as in the P2P registration approach~\cite{yu2021cofinet}. IR curves in Fig.~\ref{fig_IR} indicate that our method achieves a higher proportion of correct matches among the established correspondences, which leads to better registration results. Fig. \ref{fig_4} intuitively illustrates that our CoFiI2P provides a much cleaner correspondence set than baseline method.

\vspace{-0.4cm}
\subsection{Ablation Studies and Analysis}
In this part, we analyze three crucial factors in our CoFiI2P: I2P transformer module, coarse-to-fine matching scheme, and point cloud density. We conduct ablation studies on the KITTI Odometry~\cite{geiger2012we} dataset to prove the effectiveness of each module and review the influence of point cloud density. We train the ablation models for 25 epochs as in the experimental part, and all other settings remain the same. We report global RRE and RTE as evaluation metrics and no thresholds are used to reject false registration scenes.

\subsubsection{Analysis of the I2P Transformer}
I2P transformer~\cite{dosovitskiy2020image} with self-attention module and cross-attention module is crucial to image-to-point cloud alignment at the global level. In this part, we conduct ablation studies to assess the effectiveness of the I2P transformer. We train the CoFiI2P without any attention module as baseline. Then, the self-attention modules and cross-attention modules are added on the coarse level respectively. As shown in Table \ref{tab:table3},  performance  of the baseline method drops significantly without any attention module. Besides, the self-attention module reduces the RRE to $1.78^{\circ}$ and the RTE about $0.41m$, and the cross-attention module reduces the RRE to $1.74^{\circ}$ and the RTE to $0.43m$ respectively. With both the self-attention and cross-attention modules, the CoFiI2P achieves the smallest registration error. Moreover, with both self-attention and cross-attention modules, the variance reduces by a large margin, which indicates that the I2P transformer block enhances both the accuracy and robustness. 
%

\subsubsection{Analysis of the Coarse-to-fine Matching Scheme}
our CoFiI2P proposes to estimate coarse correspondences at super-point/-pixel level first and then generate fine correspondences at point-pixel level sequentially. We conduct ablation experiments on the coarse-to-fine matching scheme to demonstrate that the progressive two-stage registration operates better than the one-stage registration used in previous I2P approaches. This ablation study employs the backbone with full I2P transformer blocks as baseline and evaluates the registration accuracy with only coarse or fine matching schemes. For coarse matching only, matching pairs are established on the coarse level and remapped to the fine resolution for pose estimation. By contrast, for fine matching only, matching pairs are established on the fine level directly, without guidance of coarse level correspondences. Experiment results in Table \ref{tab:table4} show that removing either the coarse matching stage or the fine matching stage leads to higher registration error and variance. We indicate that coarse-level registration provides robust correspondences and fine-level registration provides accurate matching pairs. Combining the coarse-level and fine-level registration sequentially makes it easier to access the global optimal solution in I2P registration.

\subsubsection{Analysis of the Point Cloud Density}
Given the significant impact of point cloud density on representative learning, we conducted ablation studies to examine this influence. The results in Table \ref{tab:table5} present registration accuracy and computational complexity for various point cloud densities. The findings reveal that as point cloud density decreases, the qualitative metrics deteriorate, while higher point cloud densities correspond to a sharp increase in computational complexity. It's an intuitive observation that low-density point clouds lose local structured information, and high-density point clouds carry a heavy computational burden. As a result, we opt for a compromise and select 20480 points, striking a balance between efficiency and accuracy.

\begin{table}[t]
\centering
\caption{Ablation study on the I2P transformer blocks. \textbf{SA} denotes the self-attention module and \textbf{CA} denotes the cross-attention module.}
\label{tab:table3}
\begin{tabular}{c|cc|cc}
\hline
  \rowcolor{gray!30} Baseline         & \textbf{SA}           & \textbf{CA}           & RRE($^\circ$)            & RTE(m)           \\ \hline
  \checkmark       &              &              &   $2.65\pm4.79$   &  $0.87\pm2.23$   \\
   \checkmark      &  \checkmark  &              &  $1.78\pm2.28$   &  $0.41\pm0.79$  \\ 
  \checkmark       &              &  \checkmark  &   $1.74\pm1.50$           &       $0.43\pm0.35$            \\
  \checkmark       &  \checkmark  &  \checkmark  &  $\mathbf{1.14 \pm 0.78}$   &  $\mathbf{0.29 \pm 0.19}$   \\ \hline
\end{tabular}
\vspace{0.5cm}
\setlength{\tabcolsep}{10.8pt}
\centering
\caption{Ablation study on the coarse-to-fine matching scheme. \textbf{CM} denotes the coarse matching and \textbf{FM} denotes the fine matching. }
\label{tab:table4}
\begin{tabular}{c|cc|cc}
\hline
  \rowcolor{gray!30} Baseline         & \textbf{CM}           & \textbf{FM}           & RRE($^\circ$)            & RTE(m)            \\ \hline
   \checkmark      &  \checkmark  &              &  $1.35 \pm 1.17$   &  $0.34 \pm 0.26$  \\
  \checkmark       &              &  \checkmark  &   $1.47 \pm 1.75$            &     $0.41 \pm 0.63$               \\
  \checkmark       &  \checkmark  &  \checkmark  &  $\mathbf{1.14 \pm 0.78}$   &  $\mathbf{0.29 \pm 0.19}$   \\ \hline
\end{tabular}
\vspace{0.5cm}
\setlength{\tabcolsep}{13pt}
\centering
\caption{Ablation study on the point cloud density. }
\label{tab:table5}
\begin{tabular}{c|ccc}
\hline
  \rowcolor{gray!30} \#Points         & RRE($^\circ$)           & RTE(m)             & FLOPs            \\ \hline
   5120              &   $2.75 \pm 4.28$            &    $0.69 \pm 1.60$      & 37.42G   \\    
  10240              &     $1.69 \pm 1.43$         &      $0.39 \pm 0.30$         &        56.00G                          \\
  20480              &     $1.14\pm0.78$         &            $0.29\pm0.19$    &     93.80G                          \\
  40960              &    $1.00 \pm 0.70$  &   $ 0.26 \pm 0.17$   & 171.91G \\ \hline
\end{tabular}
\end{table}

\section{Conclusion}\label{section:6}
This letter introduces CoFiI2P, a novel network designed for image-to-point cloud (I2P) registration. The proposed coarse-to-fine matching strategy first establishes robust global correspondences and then progressively refines precise local correspondences. Furthermore, the I2P transformer with self- and cross-attention modules is introduced to enhance the global-aware ability in homogeneous and heterogeneous data. Compared with existing one-stage dense prediction and matching approaches, CoFiI2P filters out a large number of false correspondences. Extensive experiments on the KITTI Odometry~\cite{geiger2012we} and Nuscenes~\cite{caesar2020nuscenes} dataset have demonstrated the superior accuracy, efficiency, and robustness of CoFiI2P in various environments. We hope the open-sourced CoFiI2P could benefit the relevant communities. In the near future, we will extend CoFiI2P to the unsupervised I2P registration.

\small{
\bibliography{cofii2p} 
\bibliographystyle{IEEEtran}
}

\end{document}